\documentclass[sigconf]{acmart}

\usepackage{amsmath}
\usepackage{subcaption}
\usepackage{footnote}
\usepackage{csquotes}
\usepackage{ragged2e}
\usepackage{graphicx}
\usepackage{tabularx}
\usepackage{tablefootnote}
\usepackage{url}
\usepackage[normalem]{ulem}
\usepackage{multirow}
\useunder{\uline}{\ul}{}

\DeclareMathOperator*{\argmax}{arg\,max}
\AtBeginDocument{%
  \providecommand\BibTeX{{%
    \normalfont B\kern-0.5em{\scshape i\kern-0.25em b}\kern-0.8em\TeX}}}

\def \hfillx {\hspace*{-\textwidth} \hfill}
\copyrightyear{2020}
\acmYear{2020}
\setcopyright{acmcopyright}
\acmConference[CoDS COMAD 2020]{7th ACM IKDD CoDS and 25th COMAD}{January 5--7, 2020}{Hyderabad, India}
\acmBooktitle{7th ACM IKDD CoDS and 25th COMAD (CoDS COMAD 2020), January 5--7, 2020, Hyderabad, India}
\acmPrice{15.00}
\acmDOI{10.1145/3371158.3371165}
\acmISBN{978-1-4503-7738-6/20/01}

\begin{document}

\title{A Unified System for Aggression Identification in English Code-Mixed and Uni-Lingual Texts}


\author{Anant Khandelwal}
\email{anant.iitd.2085@gmail.com}
\affiliation{M.Tech., 
  \institution{IIT Delhi}
%  \streetaddress{P.O. Box 1212}
  \city{Bangalore}
  \state{India}
}

\author{Niraj Kumar}
\email{nirajrkumar@gmail.com}
\affiliation{Conduent Labs, PhD, 
  \institution{IIIT-Hyderabad}
%  \streetaddress{P.O. Box 1212}
  \city{Bangalore}
  \state{India}
}
\renewcommand{\shortauthors}{Anant Khandelwal and Niraj Kumar}

\begin{abstract}
  Wide usage of social media platforms has increased the risk of aggression, which results in mental stress and affects the lives of people negatively like psychological agony, fighting behavior, and disrespect to others. Majority of such conversations contains code-mixed languages\cite{kumar2018aggression}. Additionally, the way used to express thought or communication style also changes from one social media platform to another platform (e.g., communication styles are different in twitter and Facebook). These all have increased the complexity of the problem. To solve these problems, we have introduced a unified and robust multi-modal deep learning architecture which works for English code-mixed dataset and uni-lingual English dataset both. The devised system, uses psycho-linguistic features and very basic linguistic features. Our multi-modal deep learning architecture contains, Deep Pyramid CNN, Pooled BiLSTM, and Disconnected RNN(with Glove and FastText embedding, both). Finally, the system takes the decision based on model averaging. We evaluated our system on English Code-Mixed TRAC\footnote{https://sites.google.com/view/trac1/home} 2018 dataset and uni-lingual English dataset obtained from Kaggle\footnote{https://www.kaggle.com/dataturks/dataset-for-detection-of-cybertrolls}. Experimental results show that our proposed system outperforms all the previous approaches on English code-mixed dataset and uni-lingual English dataset.  
\end{abstract}



\maketitle

\raggedbottom

\section{Introduction}
The exponential increase of interactions on the various social media platforms has generated the huge amount of data on social media platforms like Facebook and Twitter, etc. These interactions resulted not only positive effect but also negative effect over billions of people owing to the fact that there are lots of aggressive comments (like hate, anger, and bullying). These cause not only mental and psychological stress but also account deactivation and even suicide\cite{hinduja2010bullying}. In this paper we concentrate on problems related to aggressiveness.
\newline The fine-grained definition of the aggressiveness/aggression identification is provided by the organizers of TRAC-2018 \cite{kumar2018aggression, kumar2018trac}. They have classified the aggressiveness into three labels (Overtly aggressive(OAG), Covertly aggressive(CAG), Non-aggressive(NAG)). The detailed description for each of the three labels is described as follows:
\begin{enumerate}
\item \textbf{Overtly Aggressive(OAG)} - This type of aggression shows direct verbal attack pointing to the particular individual or group. For example, "\textit{Well said sonu..you have courage to stand against dadagiri of Muslims}".
\item \textbf{Covertly Aggressive(CAG)} - This type of aggression the attack is not direct but hidden, subtle and more indirect while being stated politely most of the times. For example, "\textit{Dear India,  stop playing with the emotions of your people for votes.}"
\item \textbf{Non-Aggressive(NAG)} - Generally these type of text lack any kind of aggression it is basically used to state facts, wishing on occasions and polite and supportive.
\end{enumerate}
The additional discussion on aggressiveness task can be found in Kaggle task \footnote{\url{https://www.kaggle.com/dataturks/dataset-for-detection-of-cybertrolls}}, which just divided the task into two classes - i.e., presence or absence of aggression in tweets.
\newline
The informal setting/environment of social media often encourage multilingual speakers to switch back and forth between languages when speaking or writing. These all resulted in \enquote{code-mixing} and \enquote{code-switching}. Code-mixing refers to the use of linguistic units from different languages in a single utterance or sentence, whereas code-switching refers to the co-occurrence of speech extracts belonging to two different grammatical systems\cite{gumperz1982discourse}. This language interchange makes the grammar more complex and thus it becomes tough to handle it by traditional algorithms. Thus the presence of high percentage of code-mixed content in social media text has increased the complexity of the aggression detection task. For example, the dataset provided by the organizers of TRAC-2018 \cite{kumar2018aggression, kumar2018trac} is actually a code-mixed dataset.
\newline
The massive increase of the social media data rendered the manual methods of content moderation difficult and costly. Machine Learning and Deep Learning methods to identify such phenomena have attracted more attention to the research community in recent years\cite{kumar2018benchmarking}.
\newline
Based on the current context, we can divide the problem into three sub-problems: (a) detection of aggression levels, (b) handling code-mixed data and (c) handling styles (due to differences in social media platforms and text entry rules/restrictions).
\newline
A lot of the previous approaches\cite{kumar2018proceedings} have used an ensemble model for the task. For example, some of them uses ensemble of statistical models\cite{arroyo2018cyberbullying, fortuna2018merging, samghabadi2018ritual, orasan2018aggressive} some used ensemble of statistical and deep learning models\cite{risch2018aggression, tommasel2018textual, ramiandrisoa2018irit} some used  ensemble of deep learning models \cite{madisetty2018aggression}. There are approaches which proposed unified architecture based on deep learning\cite{aroyehun2018aggression, golem2018combining, orabi2018cyber, raiyani2018fully, nikhil2018lstms, galery2018aggression} while some proposed unified statistical model\cite{fortuna2018merging}. Additionally, there are some approaches uses data augmentation either through translation or labeling external data to make the model generalize across domains\cite{aroyehun2018aggression, risch2018aggression, fortuna2018merging}. 
\newline
Most of the above-discussed systems either shows high performance on (a) Twitter dataset or (b) Facebook dataset (given in the TRAC-2018), but not on both English code-mixed datasets. This may be due to the text style or level of complexities of both datasets. So, we concentrated to develop a robust system for English code-mixed texts, and uni-lingual texts, which can also handle different writing styles. Our approach is based on three main ideas: 
\begin{itemize}
	\item \textbf{Deep-Text Learning}. The goal is to learn long range associations, dependencies between regions of text, N-grams, key-patterns, topical information, and sequential dependencies.
	\item \textbf{Exploiting psycho-linguistic features with basic linguistic features as meta-data}. The main aim is to minimize the direct dependencies on in-depth grammatical structure of the language (i.e., to support code-mixed data). We have also included emoticons, and punctuation features with it. We use the term "NLP Features" to represent it in the entire paper.
	\item \textbf{Dual embedding based on FastText and Glove}. This dual embedding helps in high vocabulary coverage and to capture the rare and partially incorrect words in the text (specially by FastText \cite{mikolov2018advances}).  
\end{itemize}
Our "Deep-text architecture" uses model averaging strategy with three different deep learning architectures. Model averaging belongs to the family of ensemble learning techniques that uses multiple models for the same problem and combines their predictions to produce a more reliable and consistent prediction accuracy \cite{ju2018relative}. This is the simplest form of weighted average ensemble based prediction\cite{pawlikowski2019weighted} where, each ensemble member contribute equally to predictions. Specifically in our case, three different models have been used. The following contains the intuition behind the selection of these three models:
\begin{enumerate}
    \item \textbf{Deep Pyramid CNN} \cite{johnson2017deep} being deeper helps to learn long range associations between temporal regions of text using two-view embeddings.
    \item \textbf{Disconnected RNN} \cite{wang2018disconnected} is very helpful in encoding the sequential information with temporal key patterns in the text.
    \item \textbf{Pooled BiLSTM} In this architecture the last hidden state of BiLSTM is concatenated with mean and max-pooled representation of the hidden states obtained over all the time steps of Bi-LSTM. The idea of using mean and max pooling layers together is taken from \cite{RuderH18} to avoid the loss of information in longer sequences of texts and max-pooling is taken to capture the topical information\cite{shen2014latent}. 
    \item \textbf{NLP Features} In each of the individual models, the NLP features are concatenated with last hidden state before the softmax classification layer as meta-data. The main aim is to provide additional information to the deep learning network.
\end{enumerate}
The intuition behind the NLP features are the following: 
\begin{itemize}
\item \textbf{Emotion Sensor Dataset} We have introduced to use of emotion sensor features, as a meta-data information. We have obtained the word  sensor dataset from Kaggle\footnote{https://www.kaggle.com/iwilldoit/emotions-sensor-data-set}. In this dataset each word is statistically classified into 7 distinct classes (Disgust, Surprise, Neutral, Anger, Sad, Happy and Fear) using Naive Bayes, based on sentences collected from twitter and blogs.
\item \textbf{Controlled Topical Signals from Empath\footnote{http://empath.stanford.edu/}}. Empath can analyse the text across 200 gold standard topics and emotions. Additionally, it uses neural embedding to draw connotation among words across more than 1.8 billion words. We have used only selected categories like violence, hate, anger, aggression, social media and dispute from 200 Empath categories useful for us  unlike\cite{ramiandrisoa2018irit} which takes 194 categories. 
\item \textbf{Emoticons} frequently used on social media indicates the sense of sentence\cite{raiyani2018fully, galery2018aggression, orasan2018aggressive}.
\item \textbf{Normalized frequency of POS tags} According to \cite{ramiandrisoa2018irit, tommasel2018textual, fortuna2018merging, golem2018combining} POS Tags provide the degree of target aggressiveness. Like\cite{ramiandrisoa2018irit}, we have used only four tags (a) adjective (JJ, JJR, JJS), (b) adverb (RB, RBR, RBS), (c) verb (VB, VBD, VBG, VBN, VBP, VBZ) and (d) noun (NN, NNS, NNP, NNPS) (See \enquote{Penn-Treebank POS Tags}\footnote{https://www.ling.upenn.edu/courses/Fall\textunderscore 2003/ling001/penn\textunderscore treebank\textunderscore pos.html} for abbreviations and the full list). The main reason behind the selection of these four tags is to just identify words related to persons, activities, quality, etc, in the text.
\item \textbf{Sentiment polarity} obtained from VADER Sentiment Analysis \cite{hutto2014vader} (positive, negative and neutral) like used in \cite{golem2018combining, risch2018aggression, tommasel2018textual,fortuna2018merging}. It helps to demarcate aggressiveness with non-aggressiveness in the text.
\end{itemize}
\begin{figure*}[t]
    \centering
    \includegraphics[width=\linewidth]{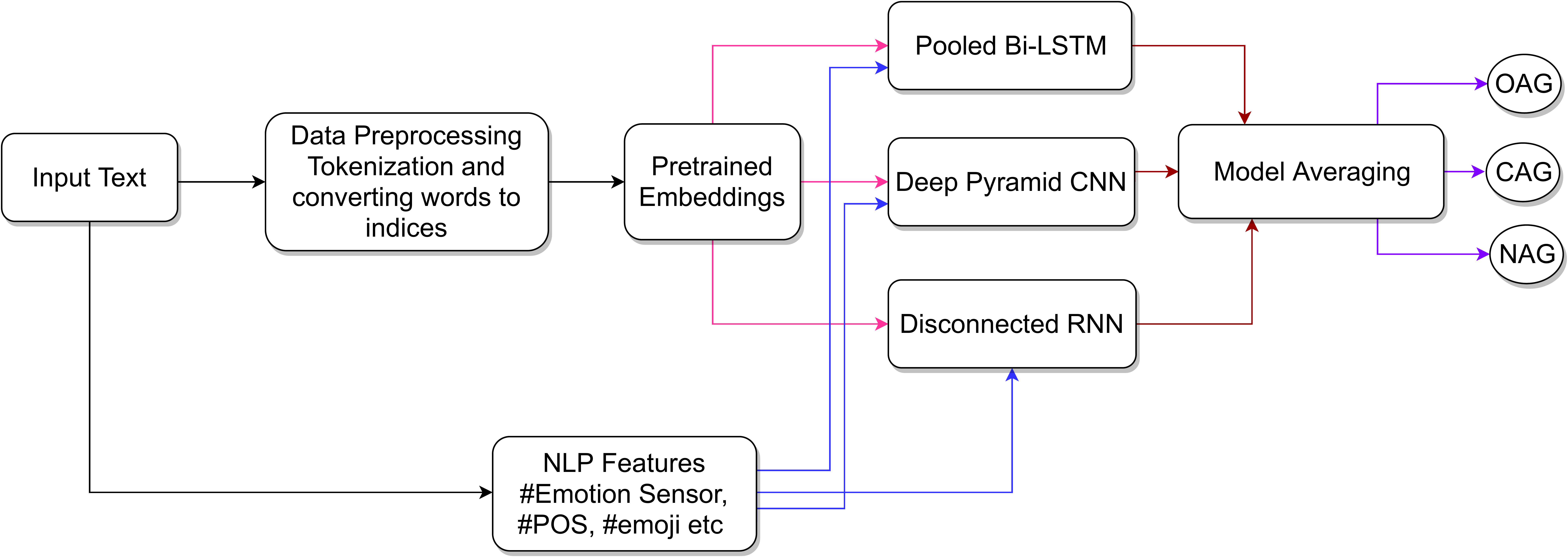}
    \caption{Block diagram of the proposed system}
    \label{fig:block}
\end{figure*}
The block diagram of the proposed system is shown in Figure \ref{fig:block}. The proposed system does not use any data augmentation techniques like \cite{aroyehun2018aggression}, which is the top performer in TRAC (in English code-mixed Facebook data). This means the performance achieved by our system totally depends on the training dataset provided by TRAC. This also proves the effectiveness of our approach. Our system outperforms all the previous state of the art approaches used for aggression identification on English code-mixed TRAC data, while being trained only from Facebook comments the system outperforms other approaches on the additional Twitter test set. The remaining part of this paper is organized as follows: Section \ref{rel_work} is an overview of related work. Section \ref{methodology} presents the methodology and algorithmic details. Section \ref{exp} discusses the experimental evaluation of the system, and Section \ref{conclusion} concludes this paper. 
\section{Related work}\label{rel_work}
There are several works for aggression identification submitted at TRAC 2018 among them some approaches use the ensemble of multiple statistical models\cite{arroyo2018cyberbullying, fortuna2018merging, samghabadi2018ritual, orasan2018aggressive}. Similarly, some of the models like\cite{risch2018aggression, tommasel2018textual, ramiandrisoa2018irit} have used ensemble of statistical and deep learning models. In these models the statistical part of the model uses additional features from text analysis like parts-of-speech tags, punctuation, emotion, emoticon etc. Model like: \cite{madisetty2018aggression} has used the ensemble of deep learning models based on majority voting. 
\newline Some other models like: \cite{majumder2018filtering, ramiandrisoa2018irit, orasan2018aggressive} have used different models for Facebook and twitter. While approaches like:\cite{aroyehun2018aggression, golem2018combining, orabi2018cyber, raiyani2018fully, nikhil2018lstms, galery2018aggression} have proposed unified architecture based on deep learning. Systems like\cite{aroyehun2018aggression, risch2018aggression, fortuna2018merging} have used data augmentation either through translation or labelling external data to make the model generalize across domains. While \cite{fortuna2018merging} has proposed a unified statistical model. 
\newline Among approaches like\cite{arroyo2018cyberbullying} extracted features from TF-IDF of character n-grams while\cite{majumder2018filtering} uses LSTM with pre-trained embeddings from FastText. \cite{golem2018combining}  have used the BiLSTM based model and the SVM metaclassifier model for the Facebook and Twitter test sets, respectively. While \cite{madisetty2018aggression} tried ensembling of CNN, LSTM, and BILSTM. 
\newline Some approaches like:\cite{ramiandrisoa2018irit} has used emotions frequency as one of the features, while some others use sentiment emotion as feature\cite{tommasel2018textual}. Also,\cite{raiyani2018fully, galery2018aggression} have converted emoticons to their description. \cite{orasan2018aggressive} have used TF-IDF of emoticons per-class as one of the features. Compared to all these approaches, we have concentrated to capture multiple linguistic/pattern based relations, key-terms and key-patters (with their association in text) through a combination of deep learning architectures with model averaging. We have also used NLP features as additional features with our deep learning architecture, obtained from psycho-linguistic and basic linguistic features.
\section{Methodology}\label{methodology}
In this section, we describe our system architecture for aggressiveness classifier. In section \ref{datap} we describe data preprocessing applied on the input text before feeding it to each of the classification models. Section \ref{nlp} describes the computation of NLP features. In Sections \ref{dpcnn}, \ref{drnn} and \ref{pbilm} we have described the architecture of different deep learning models like Deep Pyramid CNN, Disconnected RNN and Pooled BiLSTM respectively.  Finally, in Section \ref{avgm}, we describe model averaging based classification model which combines the prediction probabilities from three deep learninig architectures discussed above.  (see Figure \ref{fig:block}. for block diagram of system architecture).

\subsection{Data Preprocessing}\label{datap}

We consider the text to be well formatted before applying the text to the embedding layer. First, we detect non-English text(which are few) and translate all of them to English using Google Translate\footnote{https://pypi.org/project/googletrans/}. Still, there is some code mixed words like "mc", "bc" and other English abbreviations and spelling errors like "nd" in place of "and", "u" in place of "you" causes deep learning model to confuse with sentences of the same meaning. We follow the strategy of preprocessor as in\cite{raiyani2018fully} to normalize the abbreviations and remove spelling errors, URLs and punctuation marks, converting emojis to their description.
\begin{table*}[htb]
\begin{tabular}{|l|p{12cm}|l|}
\hline
Feature Name                                                        & Description                                                                                                                                                                                                                                                                                                                                                                                                                                                                                                                                                                    & Feature Count \\ \hline
\begin{tabular}[c]{@{}l@{}}Emotion Sensor Feature
\end{tabular}  & \begin{tabular}[c]{@{}p{12cm}@{}}
Emotion Sensor Features are developed by classifying the words statistically using Naive Bayes algorithm into 7 different categories (Disgust, Surprise, Neutral, Anger, Sad, Happy and Fear) using sentences collected from twitter or blogs.\end{tabular} & 7             \\ \hline
Parts-of-Speech(POS)                                                & \begin{tabular}[c]{@{}p{12cm}@{}}  Like\cite{ramiandrisoa2018irit} we have used the normalized frequencies of noun, adjective, verb, and adverb which serve as a rich feature for exaggerating type of aggressiveness in the text. We used the Spacy\footnotemark \textrm{ }POS tagger.\end{tabular} & 4             \\ \hline
Punctuation                                                         & \begin{tabular}[c]{@{}p{12cm}@{}}Like in \enquote{\textit{Communist parties killed lacks of opponents in WB in 35 years ruling????? ?}} the presence of multiple question marks linked with the aggressiveness content in the text. Similarly "!" also has the same effect. We have used the count of \enquote{!} \& \enquote{?} in the text as a feature.\end{tabular}                                                                                                                                                                                                                                                                                                        & 1             \\ \hline
Sentiment analysis                                                  & \begin{tabular}[c]{@{}p{12cm}@{}}The percentage of positive, negative, and neutral can indicate the amount of aggressiveness in the text. We have used VADER Sentiment Analysis\cite{hutto2014vader} to extract the weight for positive, negative  \& neutral class.\end{tabular}                                                                                                                                                                                                                                                                                                                                  & 3             \\ \hline
Topic Signals from text                                             & \begin{tabular}[c]{@{}p{12cm}@{}}We have used open source library Empath{[}2{]} introduced in \cite{fast2016empath} which has categories highly correlated to LIWC\cite{pennebaker2007linguistic}. Although it has a rich number of categories we particularly identified selected categories useful for our case these are aggressiveness, violence, hate, anger, dispute \& social\_media. We have used the normalized weight of each of these categories as separate features.\end{tabular} & 6             \\ \hline
\begin{tabular}[c]{@{}l@{}}TF-IDF Emoticon Feature\end{tabular} & \begin{tabular}[c]{@{}p{12cm}@{}}Since emoticons are widely used to convey the meaning on social media platform analyzing the data emoticons are an important feature for classifying aggressive behavior and tf-idf feature for each class is calculated as in \cite{orasan2018aggressive}\end{tabular}                                                                                                                                                                                                                                                                              & 3             \\ \hline
\begin{tabular}[c]{@{}l@{}}Total Features\end{tabular}          &                                                                                                                                                                                                                                                                                                                                                                                                                                                                                                                                                                                & 24            \\ \hline
\end{tabular}
\caption{Details of NLP features}
\label{tab:feat}
\vspace{-4mm}
\end{table*}
\footnotetext{https://spacy.io/usage/linguistic-features\#pos-tagging}
\raggedbottom
\subsection{NLP Features}\label{nlp}
We have identified a novel combination of features which are highly effective in aggression classification when applied in addition to the features obtained from the deep learning classifier at the classification layer. We have introduced two new features in addition to the previously available features. The first one is the Emotion Sensor Feature\footnote{https://www.kaggle.com/iwilldoit/emotions-sensor-data-set}  which use a statistical model to classify the words into 7 different classes based on the sentences obtained from twitter and blogs which contain total 1,185,540 words. The second one is the collection of selected topical signal from text collected using Empath\footnote{http://empath.stanford.edu/} (see Table 1.).\newline
Different from previous approaches\cite{samghabadi2018ritual, ramiandrisoa2018irit} where \cite{ramiandrisoa2018irit} have used Emotion features in the form of frequency while \cite{samghabadi2018ritual} have used emotion feature vector obtained from LIWC 2007\cite{pennebaker2007linguistic}. Unlike\cite{ramiandrisoa2018irit} we have used only 6 topical signals from Emapth\cite{fast2016empath}. We have borrowed the idea of using other features like punctuation features and parts-of-speech tags from \cite{ramiandrisoa2018irit}. The Table 1. lists and describes features, tools used to obtain them and the number of features resulted from each type.
\raggedbottom
\begin{figure}[h]
    \centering
    \includegraphics[width=\linewidth]{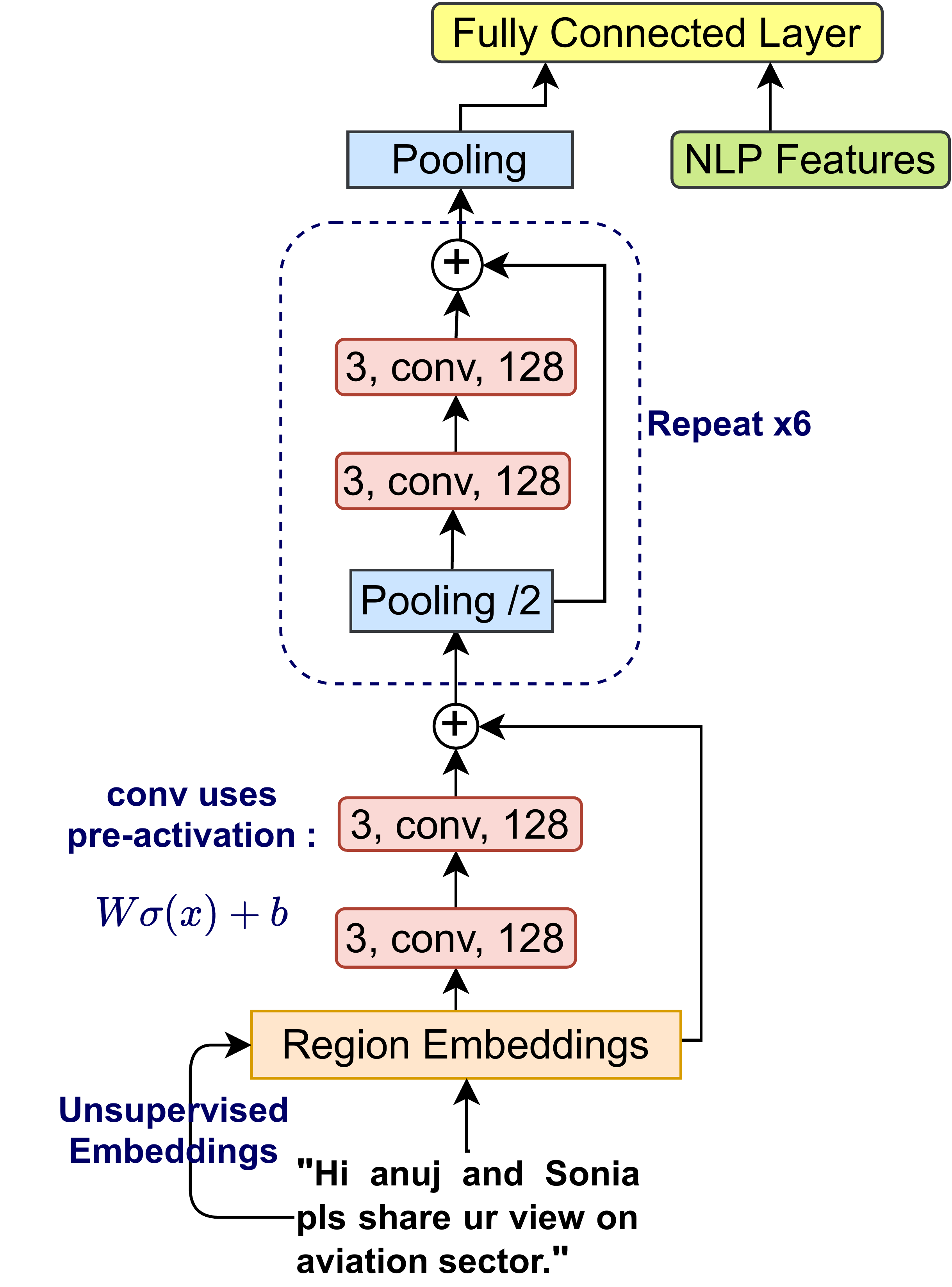}
    \caption{DPCNN}
    \label{fig:dpcnn}
\end{figure}
\subsection{Deep Pyramid CNN(DPCNN)}\label{dpcnn}
Since it has been proved that CNNs are great feature extractors for text classification\cite{yin2017comparative, kim2014convolutional, johnson2014effective, johnson2015semi, johnson2016convolutional, johnson2017deep} while deeper networks(whether RNNs or CNN's) has been proven for learning long-range association like deeper character level CNN's\cite{zhang2015character, conneau2016very}, and complex combination of RNN and CNN\cite{peng2019hierarchical, shen2019towards, jiao2019higru, yu2018mattnet, yang2016hierarchical}. Deep Pyramid CNN (DPCNN)\cite{johnson2017deep} has 15 layers of word-level CNN's and contains similar pre-activation as proposed in improved Resnet\cite{he2016identity}. DPCNN outperforms the 32-layer character CNN\cite{conneau2016very} and Hierarchical attention networks\cite{yang2016hierarchical} it has added advantage that due to its pyramid structure it does not require dimension matching in shortcut connections defined as z +  h(z) as in\cite{he2016identity} where h(z) represents the skipped layers essentially contains two convolutional layers with pre-activation. It uses enhanced region embedding which consumes pre-trained embeddings (in our case it is FastText+Glove based dual embedding).
\newline \textbf{Enhanced Region Embedding.} The current DPCNN\cite{johnson2017deep}, uses two view type enhanced region embedding. For the text categorization, it defines a region of text as view-1 and its adjacent regions as view-2. Then using unlabeled data, it trains a neural network of one hidden layer with an artificial task of predicting view-2 from view-1. The obtained hidden layer, which is an embedding function that takes view-1 as input, serves as an unsupervised embedding function in the model for text categorization. The detailed architecture has been shown in Figure \ref{fig:dpcnn}. 
\newline Let each word input $x_j \in R^d$ be the d-dimensional vector for the $j^{th}$ word $w_{j}$ and the sentence $s_i$ contains sequence of $n$ words $\{w_{1},w_{2},w_{3},......,w_{n}\}$ as shown in Figure \ref{fig:dpcnn}.
In comparision to conventional convolution layer, DPCNN proposes to use pre-activation, thus essentially the convolutional layer of DPCNN is $\textbf{W}\sigma(\textbf{x})+\textbf{b}$, where $\textbf{W}$ and $\textbf{b}$(unique to each layer) are the weights matrix and bias respectively, we use $\sigma$ as PReLU\cite{he2015delving}. During implementation we use kernel size of 3(represented by $\textbf{x}$ to denote the small overlapping regions of text.), The number of filters(number of feature maps denoted by the number of rows of $\textbf{W}$) is 128 as depicted in Figure \ref{fig:dpcnn}. With the number of filters same in each convolution layer and max-pooling with stride 2 makes the computation time halved, and doubles the net coverage of convolution kernel. Thus the deeper layers cause to learn long-range associations between regions of text. Let's say $h_{dpcnn} \in R^{p_1}$ be the hidden state obtained from DPCNN just before the classification layer and $f_{nlp}  \in R^{24}$  be the NLP features computed from the text. Lets $z_1 \in R^{p_1 + 24}$ be another hidden state obtained as
\begin{equation}
    z_1 = h_{dpcnn} \oplus f_{nlp}
\end{equation}
where, $\oplus$ denotes concatenation. The vector $z_1$ obtained, then fed to the fully connected layer with softmax activation. Let $y_{i1}^*$ be the softmax probabilities, specifically for class label $k$ is given as:
\begin{equation}
    y_{i1,k}^{*} = p(y_i=k|s_i) = softmax(W_{dpcnn}^T z_1 + b_{dpcnn})[k] \textrm{ } \forall k \in [1...K]
    \label{softdpcnn}
\end{equation}
where $K$ is the number of classes, $W_{dpcnn}$ and $b_{dpcnn}$ are the weight matrix and bias respectively.
\begin{figure}[h]
    \centering
    \includegraphics[width=\linewidth]{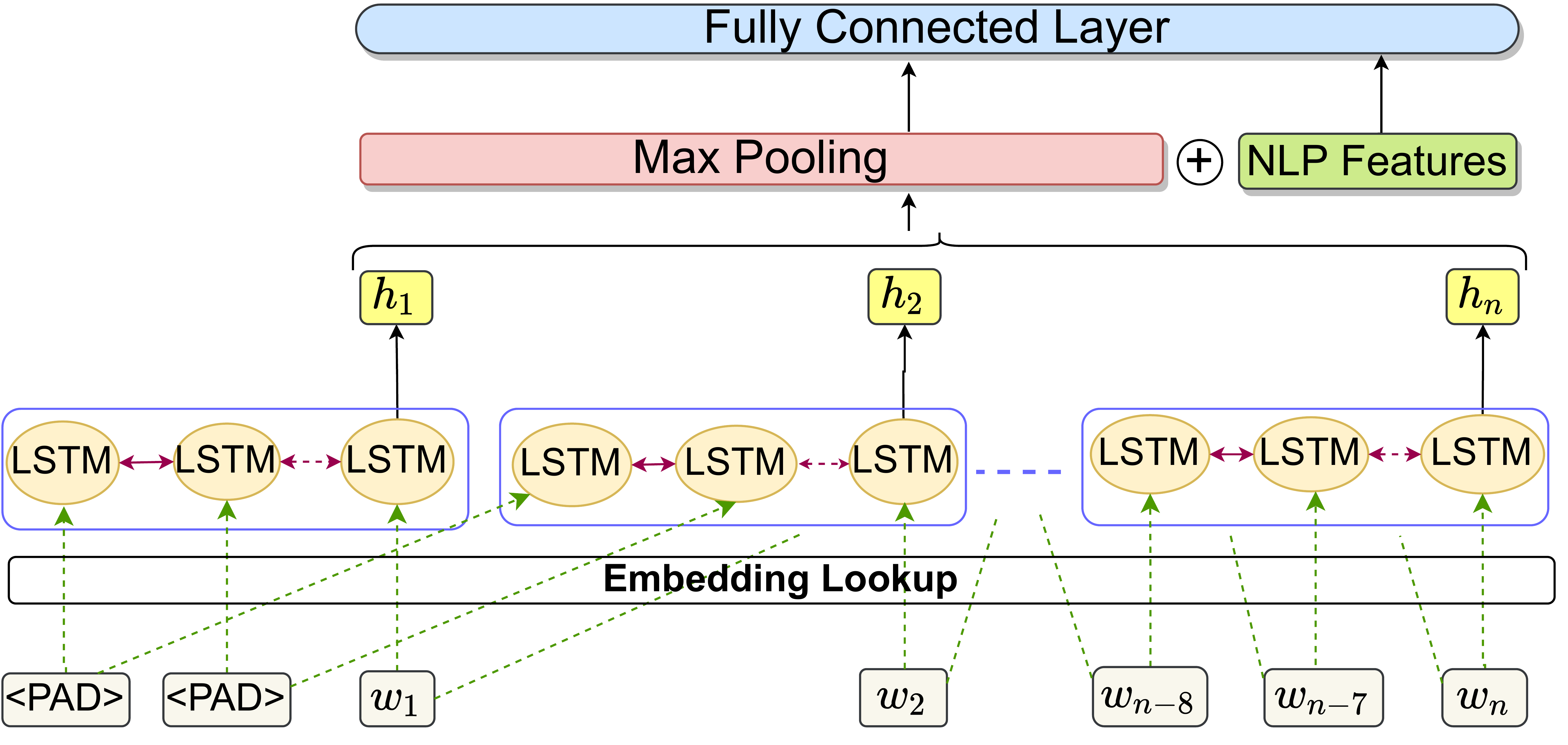}
    \caption{DRNN}
    \label{fig:drnn}
    \vspace{-2mm}
\end{figure}
\subsection{Disconnected RNN(DRNN)}\label{drnn}
Given a sequence $s_i = [x_{1}, x_{2}, x_{3},....x_{n}]$ where $x_{j} \in R^d$ represents the d-dimensional word vector for word $w_{j}$ and $n$ is the length of input text applied to a variant of RNN called Long Short-Term Memory (LSTM)\cite{hochreiter1997long} as shown in Figure \ref{fig:drnn}. It is widely used for sequential modelling with long-term dependencies.  For sequence modelling it keeps on updating the memory cell with current input using an adaptive gating mechanism. At time step $t$ the memory $c_t$ and the hidden state $h_t$ are updated as follows:
\begin{equation}
    \begin{bmatrix}
i_t\\ 
f_t\\ 
o_t\\ 
\hat{c}_t
\end{bmatrix} = \begin{bmatrix}
\sigma\\ 
\sigma\\ 
\sigma\\ 
tanh
\end{bmatrix} = W . [h_{t-1}, x_{t}]
\end{equation}
\begin{equation}
    c_t = f_t \odot c_{t-1} + i_t \odot \hat{c}_t 
\end{equation}
where $\hat{c}_t$ is the current cell state obtained from current input $x_t$ and previous hidden state $h_{t-1}$, $i_t$, $f_t$ and $o_t$ are the activation corresponding to input gate, forget gate and output gate respectively, $\sigma$ denotes the logistic sigmoid function and $\odot$ denotes the element-wise multiplication.
Hence the hidden state representation at time step $t$ depends on all the previous input vectors given as
\begin{equation}
    h_t = LSTM(x_{t}, ...., x_{2}, x_{1}) \textrm{ }\forall t\in [1...n]
    \label{shlstm}
\end{equation}
Specifically we have used Bi-directional LSTM \cite{hochreiter1997long} to capture both past and future context. It provides $h_t$ from both directions(forward \& backward). The forward LSTM takes the natural order of words from $x_{1}$ to $x_{n}$ to obtain $\overrightarrow{h_t}$, while backward-LSTM $x_{n}$ to $x_{1}$ to obtain $\overleftarrow{h_t}$. then $h_t$ is calculated as
\begin{equation}
    h_t = \overrightarrow{h_t} \oplus \overleftarrow{h_t} \in R^{2L}
\end{equation}
where $\oplus$ is the concatenation and $L$ is the size for one-directional LSTM. Therefore we denote the hidden state in equation \ref{shlstm} with BiLSTM as 
\begin{equation}
    h_t = BiLSTM(x_{t}, ,......x_{2}, x_{1}) \in R^{2L}\textrm{ } \forall t\in [1..n]
\end{equation}
To avoid handling of long sequence and to capture local information for each word we define the window size $k$ for each word such that the BiLSTM only sees the the previous $k-1$ words with the current word, where $k$ is a hyperparameter\cite{wang2018disconnected}. We use padding <PAD> to make the slices of fixed size k(as shown in Figure \ref{fig:drnn}). It provides each hidden state $h_t$ with sequence of $k$ previous words. Since the phrase of $k$ words can lie anywhere in the text it helps to model the position invariant phrase representation due to which the it identifies key phrases important for identifying particular category. In this case, the equation of $h_t$ is given as 
\begin{equation}
    h_t = BiLSTM(x_{t}, x_{t-1}, x_{t-2}, ......, x_{t-k+1}) 
\end{equation}
The output hidden vectors, $H = [h_1, h_2, h_3, ...... h_n] \in R^{n \times 2L}$ are converted to fixed-length vector $h_{drnn} \in R^{2L}$ with max pooling over time:  
\begin{equation}
    h_{drnn}^{l} = \underset{t}{max}\textrm{ }h_{t}^{l}, t \in [1,...n], \forall l \in [1,.....2L] 
\end{equation}
Let's say $f_{nlp} \in R^{24}$ be the NLP features computed from the text. Let's $z_2 \in R^{2L + 24}$ be another hidden state obtained as
\begin{equation}
    z_2 = h_{drnn} \oplus f_{nlp}   
\end{equation}
where $\oplus$ denotes concatenation. The vector $z_2$ obtained, then fed to the fully connected layer with softmax activation. Let $y_{i2}^*$ be the softmax probabilities, specifically for class label $k$ is given as:
\begin{equation}
    y_{i2,k}^{*} = p(y_i = k |s_i) = softmax(W_{drnn}^T z_2 + b_{drnn})[k]\textrm{ } \forall k \in [1...K]
    \label{softdrnn}
\end{equation}
where $K$ is the number of classes, $W_{drnn}$ is the weight matrix, and
$b_{drnn}$ is the bias.
\begin{figure}[t]
    \centering
    \includegraphics[width=\columnwidth]{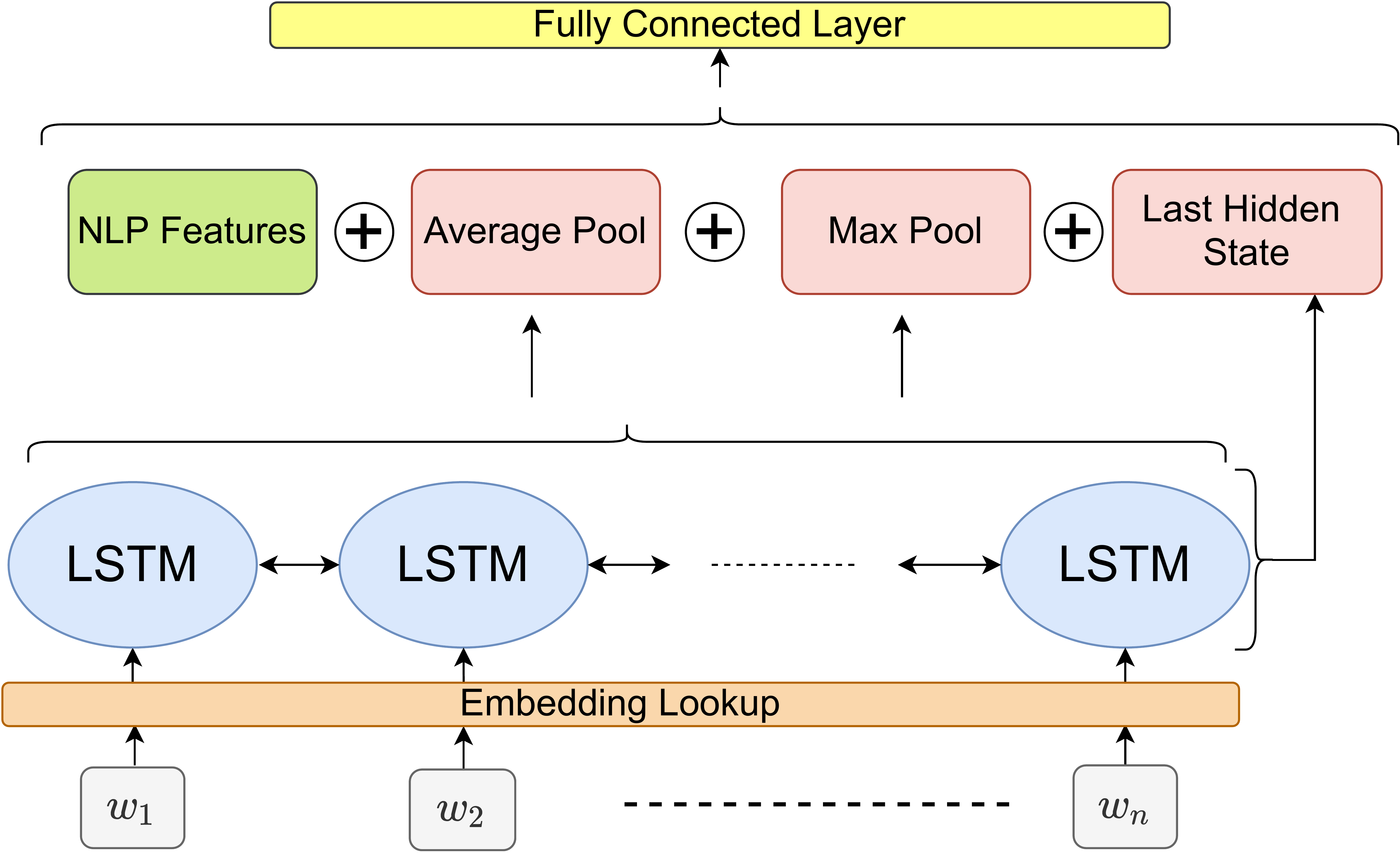}
    \caption{Pooled BiLSTM}
    \label{fig:pooled}
    \vspace{-4mm}
\end{figure}
\raggedbottom
\subsection{Pooled BiLSTM}\label{pbilm}
The architecture has been shown in Figure \ref{fig:pooled}. Given a sequence $s_i = [x_{1}, x_{2}, x_{3}, ..... x_{j}]$, where $x_j \in R^d$ is the d-dimensional word vector for word $w_j$, the hidden state obtained after BiLSTM is given as 
\begin{equation}
    h_t = BiLSTM(x_{t}, x_{t-1},.... x_{1}) \forall t \in [1...n]
\end{equation}
\raggedbottom
To avoid the loss of information because of modelling the entire sequence, we have concatenated the max-pooled($c_{max}$) and mean-pooled($c_{mean}$) representation of hidden states calculated over all time steps \cite{RuderH18}. We have also concatenated the nlp features, $f_{nlp} \in R^{24}$ the final feature vector $z_{3}$ is given as
\begin{equation}
    z_3 = h_{n} \oplus c_{max} \oplus c_{mean} \oplus f_{nlp}
\end{equation}
where $\oplus$ denotes concatenation. The final feature $z_3$ vector is fed to the fully connected layer with softmax activation. Let $y_{i3}^*$ be the softmax probablities, specifically for class label $k$ given as:
\begin{equation}
    y_{i3,k}^{*} = p(y_i = k |s_i) = softmax(W_{bilstm}^T z_3 + b_{bilstm})[k]\textrm{ } \forall k \in [1...K]
    \label{softpblm}
\end{equation}
where $K$ is the number of classes and $W_{bilstm}$ and $b_{bilstm}$ are the weight matrix and bias respectively.
\raggedbottom
\subsection{Classification Model}\label{avgm}
According to deep learning literature \cite{simonyan2014very,szegedy2015going,he2016deep}, unweighted averaging might be a reasonable ensemble for similar base learners of comparable performance. Now, similar to the information discussed in \cite{ju2018relative}, we can compute the model averaging (unweighted) by combining the softmax probabilities of three different classification models obtained from equations \ref{softdpcnn}, \ref{softdrnn},  \ref{softpblm}. The averaged class probabilities are computed as:
\begin{equation}
    y_{i,k}^* = \frac{y_{i1,k}^* +  y_{i2,k}^* + y_{i3,k}^*}{3} \forall k \in [1 .... K]    
\end{equation}
\begin{equation}
    \hat{y}_{i}= \argmax_k(y_{i}^*)
\end{equation} 
where K is the number of classes, and $\hat{y_i}$ is the predicted label for sentence $s_i$.
\section{Experiment and Evaluation}\label{exp}
\subsection{Dataset Description}
We have used two datasets in our experimental evaluations: (1) TRAC 2018 Dataset\footnote{https://sites.google.com/view/trac1/shared-task} and (2) Kaggle Dataset\footnote{https://www.kaggle.com/dataturks/dataset-for-detection-of-cybertrolls}.\\
\\
\textit{TRAC 2018 Dataset}: We have used the English code-mixed dataset provided by TRAC 2018. This dataset contains three labels, (a) Non-Aggressive(NAG), (b) Overtly-Aggressive (OAG) and (c) Covertly-Aggressive(CAG). The distribution of training, validation and test sets are described in Table \ref{tab:tracdataset}.\\
\\
\textit{Kaggle Dataset}: This dataset contains 20001 tweets which are manually labeled. The labels are divided into two categories (indicating presence or absence of aggression in tweets) AGG(Aggressive) or NAG(Non-Aggressive). We have used the same test split available in the baseline code\footnote{https://www.kaggle.com/alisaeidi92/a-very-simple-nlp-model-0-75-accuracy}. The distribution for each of the training and test is given in Table \ref{tab:kagdata}.
\newcolumntype{b}{X}
\newcolumntype{s}{>{\hsize=.5\hsize}X}
\begin{table}[h]
\begin{minipage}{\linewidth}
            \centering
\resizebox{\linewidth}{!}{%
\begin{tabular}{|p{1.5cm}|l|l|l|l|}
\hline
\multicolumn{1}{|c|}{\multirow{2}{*}{\textbf{Number}}} & \multicolumn{1}{c|}{\multirow{2}{*}{\textbf{Training}}} & \multicolumn{1}{c|}{\multirow{2}{*}{\textbf{Validation}}} & \multicolumn{2}{c|}{\textbf{Test}}   \\ \cline{4-5} 
\multicolumn{1}{|c|}{}                                 & \multicolumn{1}{c|}{}                                   & \multicolumn{1}{c|}{}                                     & \textbf{Facebook} & \textbf{Twitter} \\ \hline
Texts                                                  & 11999                                                   & 3001                                                      & 916               & 1257             \\ \hline
Overtly-aggressive                                     & 2708                                                    & 711                                                       & 144               & 361              \\ \hline
Covertly \textrm{     } aggressive                                    & 4240                                                    & 1057                                                      & 142               & 413              \\ \hline
Non-aggressive                                         & 5051                                                    & 1233                                                      & 630               & 483              \\ \hline
\end{tabular}%
}
\caption{TRAC 2018, Details of English Code-Mixed Dataset}
\label{tab:tracdataset}
\end{minipage}%
\hfill
\begin{minipage}{\linewidth}
\begin{tabularx}{\linewidth}{|b|s|s|}
\hline
\multicolumn{1}{|c|}{\textbf{Number}} & \multicolumn{1}{c|}{\textbf{Training}} & \multicolumn{1}{c|}{\textbf{Test}} \\ \hline
Texts & 15000 & 5001 \\ \hline
Aggressive(AGG) & 5867 & 1955 \\ \hline
Non-Aggressive(NAG) & 9133 & 3046 \\ \hline
\end{tabularx}
\caption{Kaggle, Uni-lingual(English) Dataset}
\label{tab:kagdata}
\end{minipage}
\vspace{-10mm}
\end{table}
\begin{table*}[h]
\begin{minipage}{0.49\linewidth}
            \centering
\resizebox{\textwidth}{!}{%
\begin{tabular}{|c|l|c|}
\hline
\multicolumn{1}{|c|}{\textbf{\begin{tabular}[c]{@{}c@{}}Rank\\ in\\ Trac\\ 2018\end{tabular}}} & \multicolumn{1}{c|}{\textbf{System}}           & \textbf{\begin{tabular}[c]{@{}c@{}}F1 (Weighted)\\ Facebook\end{tabular}} \\ \hline
\multicolumn{1}{|c|}{1}                                                                        & saroyehun\cite{aroyehun2018aggression}         & 0.6425                                                                    \\ \hline
\multicolumn{1}{|c|}{2}                                                                        & EBSI-LIA-UNAM\cite{arroyo2018cyberbullying} & 0.6315                                                                    \\ \hline
\multicolumn{1}{|c|}{3}                                                                        & DA-LD-Hildesheim\cite{majumder2018filtering}          & 0.6178                                                                    \\ \hline
\multicolumn{1}{|c|}{4}                                                                        & TakeLab\cite{golem2018combining}                    & 0.6161                                                                    \\ \hline
\multicolumn{1}{|c|}{5}                                                                        & sreeIN\cite{madisetty2018aggression}          & 0.6037                                                                    \\ \hline
\multicolumn{3}{|c|}{\textbf{Our Models}}                                                                                                                                                              \\ \hline
A                                                                        & DPCNN                                              &     0.6147                                                                       \\ \hline
B                                                                        & DRNN                                              &  0.6228                                                                          \\ \hline
C                                                                        & Pooled BiLSTM                                              &       0.6190                                                                     \\ \hline
D                                                                        & DPCNN + NLP Features                                 &   0.6530                                                                         \\ \hline
E                                                                        & DRNN + NLP Features                                 &    0.6201                                                                        \\ \hline
F                                                                        & Pooled BiLSTM + NLP Features                                 &  0.6671                                                                           \\ \hline
                                                                         & Model Averaging(A + B + C)                                         &   0.6298                                                                         \\ \hline
                                                                         & 
                                                  \begin{tabular}[c]{@{}l@{}}\textbf{Our proposed method}\\ \textbf{Model Averaging(D + E + F)}\end{tabular}                                      &    0.6770                                                                        \\ \hline
\end{tabular}
}
\caption{Facebook Test results(TRAC 2018 Dataset)}
\label{tab:fb}
\vspace{-2mm}
\end{minipage}
\hfillx
\begin{minipage}{0.49\textwidth}
            \resizebox{\textwidth}{!}{%
            \begin{tabular}{|c|l|c|}
\hline
\textbf{\begin{tabular}[c]{@{}c@{}}Rank\\ in\\ Trac\\ 2018\end{tabular}} & \multicolumn{1}{c|}{\textbf{System}}           & \textbf{\begin{tabular}[c]{@{}c@{}}F1 (Weighted)\\   Twitter\end{tabular}} \\ \hline
1                                                                        & vista.ue\cite{raiyani2018fully}                & 0.6008                                                                     \\ \hline
2                                                                        & Julian\cite{risch2018aggression}               & 0.5994                                                                     \\ \hline
3                                                                        & saroyehun\cite{aroyehun2018aggression}         & 0.5920                                                                     \\ \hline
4                                                                        & EBSI-LIA-UNAM\cite{arroyo2018cyberbullying} & 0.5716                                                                     \\ \hline
5                                                                        & uOttawa\cite{orabi2018cyber}                   & 0.5690                                                                     \\ \hline
\multicolumn{3}{|c|}{\textbf{Our Models}}                                                                                                                                                              \\ \hline
A                                                                        & DPCNN                                              &  0.5364                                                                          \\ \hline
B                                                                        & DRNN                                              &  0.5272                                                                          \\ \hline
C                                                                        & Pooled BiLSTM                                              &  0.5513                                                                          \\ \hline
D                                                                        & DPCNN + NLP Features                                 &  0.5529                                                                         \\ \hline
E                                                                        & DRNN + NLP Features                                 &    0.6189                                                                       \\ \hline
F                                                                        & Pooled BiLSTM + NLP Features                                 &  0.6227                                                                          \\ \hline
                                                                         & Model Averaging(A + B + C)                                         &     0.5967                                                                       \\ \hline
                                                                         & 
                                                  \begin{tabular}[l]{@{}l@{}}\textbf{Our proposed method}\\ \textbf{Model Averaging(D + E + F)}\end{tabular}                                      &     0.6480                                                                       \\ \hline
\end{tabular}
}
\caption{Twitter Test results(TRAC 2018 Dataset)}
\label{tab:tw}
\vspace{-2mm}
\end{minipage}%
\\
\begin{minipage}{\textwidth}
\resizebox{\linewidth}{!}{%
\begin{tabular}{|l|l|l|l|l|l|}
\hline
\multicolumn{1}{|c|}{\textbf{S.No.}} & \multicolumn{1}{c|}{\textbf{System}} & \multicolumn{1}{c|}{\textbf{ACCURACY}} & \textbf{\begin{tabular}[c]{@{}c@{}}PRECISION\\(Weighted) \end{tabular}} & \textbf{\begin{tabular}[c]{@{}c@{}}RECALL\\(Weighted) \end{tabular}} & \textbf{\begin{tabular}[c]{@{}c@{}}F1\\(Weighted) \end{tabular}} \\ \hline
                                     & Baseline                             & 0.7546                                 & 0.7660                                    & 0.7546                                 & 0.7380                                       \\ \hline
                                     A                                     &    DPCNN                                  &         0.8478                               &      0.8720                                   &           0.8478                           &          0.8497                                  \\ \hline
                B                     &         DRNN                             &          0.8620                              &           0.8666                             &                  0.8620                    &                  0.8630                          \\ \hline
                 C                    &  Pooled BiLSTM                                &            0.8646                            &          0.8726                               &                 0.8646                     &              0.8659                              \\ \hline
                                     
                  D                   &    DPCNN  + NLP Features                                &               0.7798                         &                          0.8372               &              0.7798                        &       0.7812                                     \\ \hline
                   E                  &         DRNN + NLP Features                             &        0.8820                                &                   0.8885                      &             0.8820                         &       0.8830                                     \\ \hline
                    F                 &  Pooled BiLSTM  + NLP Features                              &          0.8728                             &                    0.8855                   &           0.8728                        &          0.8742                                 \\ \hline
                                     &    Model Averaging(A + B + C))                                &                    0.8666                            &          0.8712                               &                 0.8666                     &              0.8676                                    \\ \hline
                                                                          &     \textbf{ Our proposed Mehtod(Model Averaging(D + E + F))}                                &            0.9016                           &                 0.9052                        &                0.9016                      &       0.9023                                     \\ \hline
\end{tabular}%
}
\caption{Results on Kaggle Test Dataset}
\label{tab:kaggle}
\end{minipage}
\vspace{-8mm}
\end{table*}
\subsection{Experimental Setup}
We have used Glove Embeddings\cite{pennington2014glove} concatenated with FastText Embeddings\cite{mikolov2018advances} in all the three classification models presented in this paper. Specifically, we used Glove pre-trained vectors obtained from Twitter corpus containing 27 billion tokens and 1.2 million vocabulary entries where each word is represented using 100-dimensional vector. In the case of FastText the word is represented using 300-dimensional vector. Also, we have applied spatial dropout\cite{tompson2015efficient} of 0.3 at embedding layer for DPCNN(in section \ref{dpcnn}) and Pooled BiLSTM(in section \ref{pbilm}). For DPCNN model(in \ref{dpcnn}) we have learnt 128-dimensional vector representation for unsupervised embeddings implicitly for task specific representation as in \cite{johnson2017deep}. Additionally, for DPCNN all the convolutional layers used 128 filters, kernel size of 3 and max-pooling stride 2. Additionally, in the case of DPCNN we have used kernel and bias regularizer of value 0.00001 for all convolutional kernels. The pre-activation function used in DPCNN is Parametric ReLU (PReLU) proposed in \cite{he2015delving} while the activation at each of the convolutional kernel is linear. 
For, DRNN(in section \ref{drnn}) we have used the window size of 8 and rest of the parameters related to LSTM units are same as given in\cite{wang2018disconnected}. For, Pooled BiLSTM(in section \ref{pbilm}) we have used LSTM hidden units size as 256. The maximum sequence length is 200 in all three models.
In each of the classification model the classification layer contains the fully connected layer with softmax activation with output size of 3 equal to number of classes in case of TRAC 2018 dataset and its 2 in case of Kaggle dataset. Training has been done using ADAM optimizer\cite{kingma2014adam} for DPCNN and RMSPROP\cite{tieleman2012lecture} for DRNN and Pooled Bi-LSTM models. All the models are trained end-to-end using softmax cross entropy loss\cite{chollet2018keras} for TRAC 2018 dataset and binary cross entropy loss\cite{chollet2018keras} for Kaggle dataset.
\newline To train our model for TRAC 2018 dataset, we merged the training and validation dataset and then used 10\% split from shuffled dataset to save the best model, for all classifiers. We have used only 20 NLP features (except TF-IDF Emoticon feature and Punctuation feature as given in Table \ref{tab:feat}) for Kaggle dataset (as these are not present in the Kaggle dataset).
\begin{figure*}[h]
        \begin{subfigure}[b]{0.5\textwidth}
                \includegraphics[width=\linewidth]{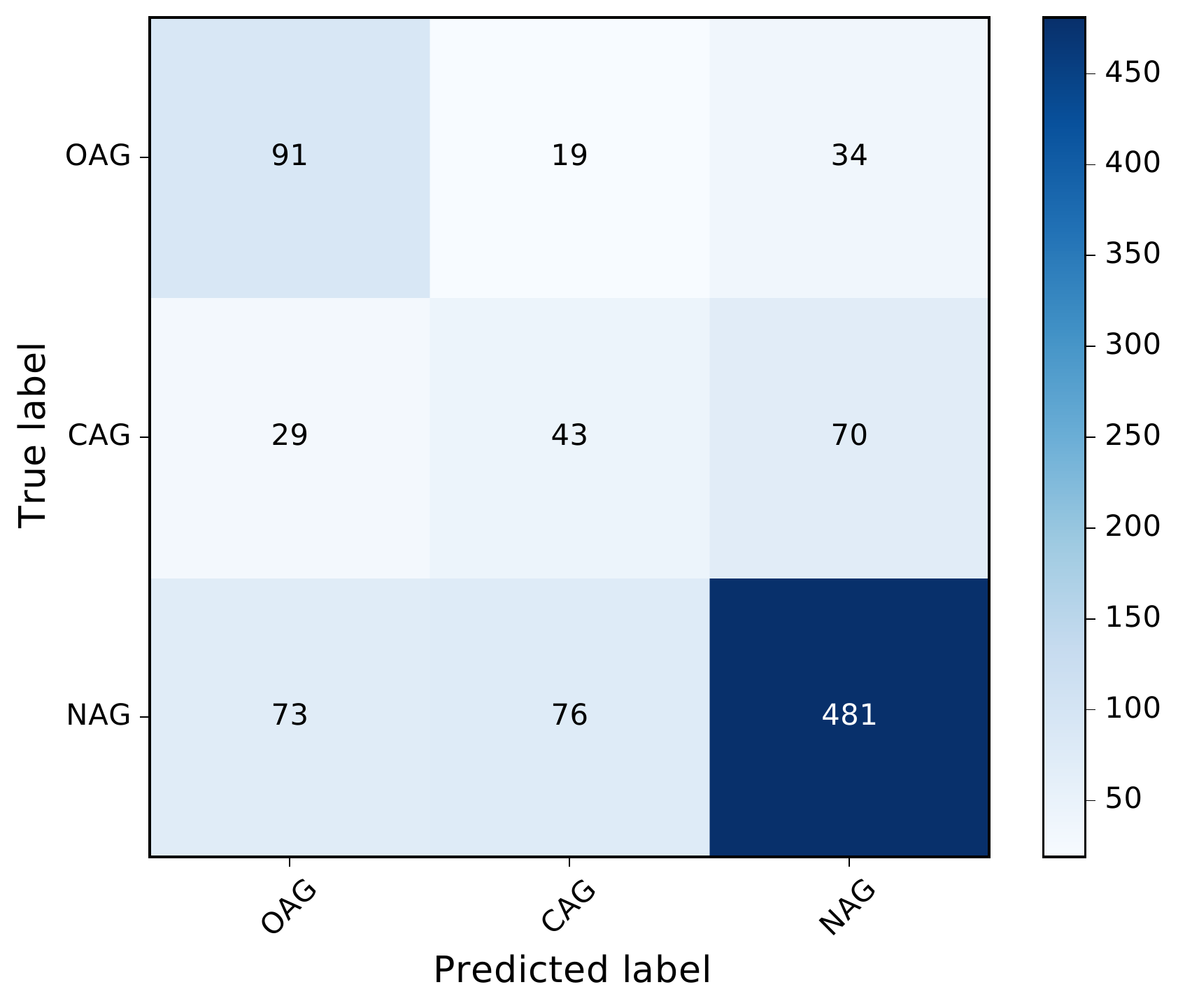}
                \caption{Confusion Matrix for Facebook Test Set(TRAC 2018)}
                \label{fig:fbconf}
 \end{subfigure}%
        \begin{subfigure}[b]{0.5\textwidth}
                \includegraphics[width=\linewidth]{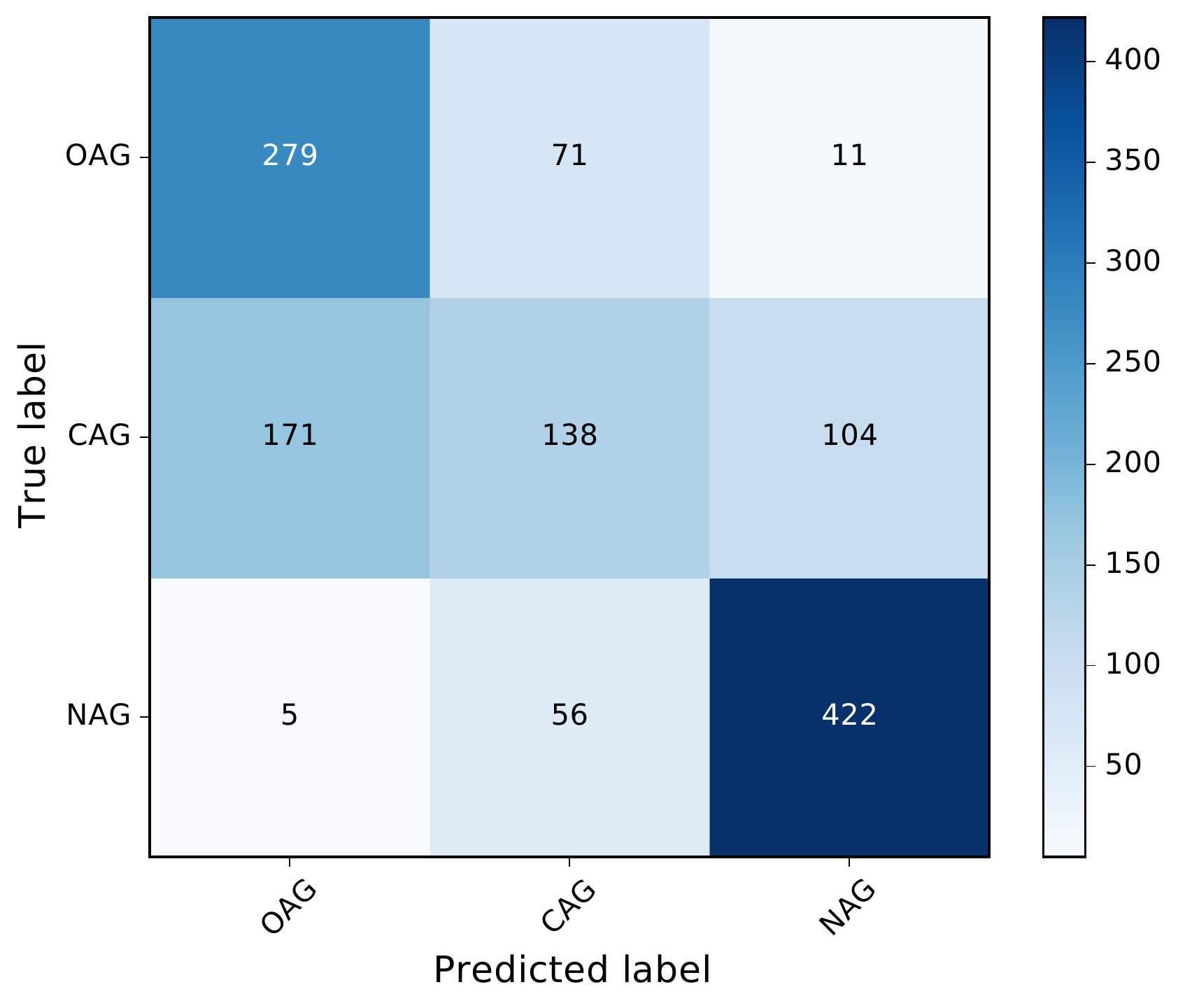}
                \caption{Confusion Matrix for Twitter Test Set(TRAC 2018)}
                \label{fig:twconf}
  \end{subfigure}%
        \\
     \begin{subfigure}[b]{0.5\linewidth}
                \includegraphics[width=\linewidth]{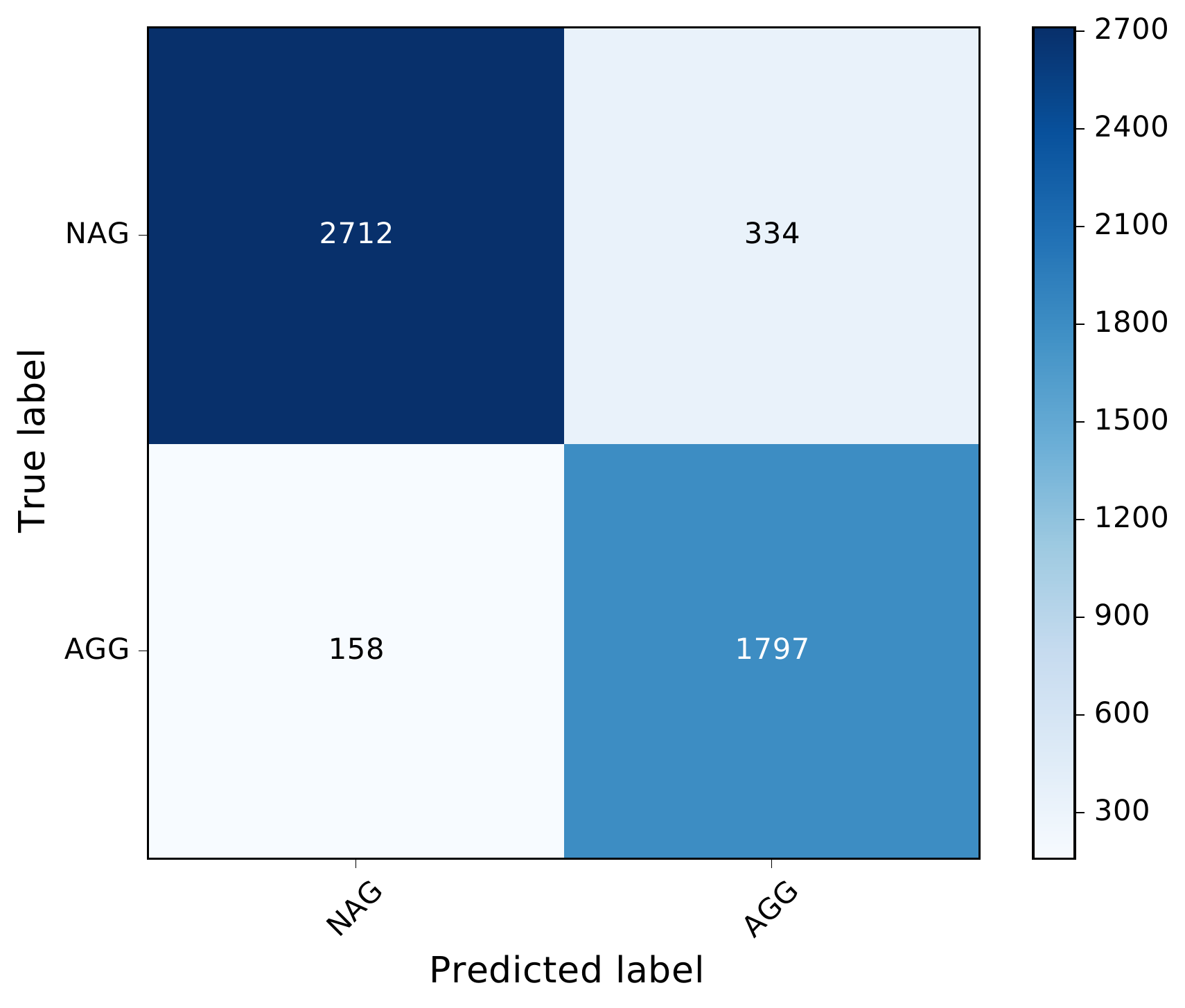}
                \caption{Confusion Matrix for Kaggle Test Set}
                \label{fig:kagconf}
      \vspace{-1em}
      \end{subfigure}%
        \caption{Confusion Matrix for Facebook, Twitter and Kaggle Datasets.}\label{fig:conf}
        \vspace{-4mm}
\end{figure*}
\subsection{Evaluation Strategy}
To compare our experimental results we have used top-5 systems from the published results of TRAC-2018\cite{kumar2018proceedings}. To compare our results on Kaggle dataset, we have used the last \& the best published result on Kaggle website as a baseline. We have conducted the separate experiments, to properly investigate the performance of (a) each of the classifiers (used in our model averaging based system), (b) impact of the NLP features on each of these classifiers and finally, (c) the performance of our proposed system. In Table \ref{tab:fb}, \ref{tab:tw} and \ref{tab:kaggle}, models, named as \textbf{DPCNN}(ref \ref{dpcnn}), \textbf{DRNN} (ref \ref{drnn}) and \textbf{Pooled BiLSTM}(ref \ref{pbilm}) are corresponding models without NLP features. Similarly, \textbf{DPCNN+NLP Features}, \textbf{DRNN + NLP Features} and \textbf{Pooled BiLSTM + NLP Features} are corresponding models with NLP features. The \textbf{Model Averaging (A+B+C)} is the ensemble of three models (i.e., model averaging of DPCNN, DRNN and Pooled BiLSTM) without NLP features. Finally, \textbf{Our Proposed Method}, which represents the model averaging of three models with NLP features. 

\subsection{Results and Discussion}\label{results}
In this paper, we have evaluated our model using weighted macro-averaged F-score. The measure is defined as in (See \cite{kumar2018proceedings, kumar2018trac}). It weights the F-score computed per class based on the class composition in the test set and then takes the average of these per-class F-score gives the final F-score. Table \ref{tab:fb}, \ref{tab:tw} and \ref{tab:kaggle}. presents the comparative experimental results for the proposed method in this paper with respect to the state-of-the-art. The top 5 models\cite{kumar2018proceedings} given in Table \ref{tab:fb} and \ref{tab:tw}. are the best performing models for Facebook and Twitter test dataset respectively on TRAC 2018. We have followed all the experimental guidelines as discussed in TRAC contest guideline paper\cite{kumar2018trac, kumar2018proceedings}. From the results given in Table \ref{tab:fb}, \ref{tab:tw} and \ref{tab:kaggle} it is clear that our proposed model shows the best performance among all of the approaches. These results also state that all the deep learning architectures with NLP features, perform better than individual corresponding deep learning architectures. This means NLP features, adds some value to the architectures, even if it is not very high.
\vspace{-1mm}
\section{Conclusion and Future Work}\label{conclusion}
In this paper, we have briefly described the approach we have taken to solve the aggressive identification on online social media texts which is very challenging since the dataset is noisy and code-mixed. We presented an ensemble of deep learning models which outperform previous approaches by sufficient margin while having the ability to generalize across domains. 
\newline In future, we will explore other methods to increase the understanding of deep learning models on group targeted text, although the categories are well defined we will look after if we further fine-tune the categories with more data. In the future, we are planning to pay attention on a generalized language model for code-mixed texts which can also handle Hindi-code-mixed and other multi-lingual code-mixed datasets (i.e., trying to reduce the dependencies on language-specific code-mixed resources). 
\bibliographystyle{ACM-Reference-Format}
\bibliography{references}

%\appendix

\end{document}